\documentclass[conference]{IEEEtran}
\IEEEoverridecommandlockouts

\usepackage{array}
\usepackage{cite}
\usepackage{amsmath,amssymb,amsfonts}
\usepackage{algorithmic}
\usepackage{graphicx}
\usepackage{textcomp}
\usepackage{xcolor}
\usepackage{booktabs}
\usepackage{listings}
\usepackage{needspace}

\usepackage{tcolorbox}
\usepackage{subcaption} 
\tcbset{colback=gray!5,colframe=black!60,boxrule=0.4pt,arc=1pt,
        left=4pt,right=4pt,top=3pt,bottom=3pt}

\def\BibTeX{{\rm B\kern-.05em{\sc i\kern-.025em b}\kern-.08em
    T\kern-.1667em\lower.7ex\hbox{E}\kern-.125emX}}
\begin{document}


\title{Implicit Framing in Obstetric Counseling Notes: A Grounded LLM Pipeline on a VBAC-Eligible Cohort}

\author{
\IEEEauthorblockN{
Baris Karacan,
Barbara Di Eugenio,
Patrick Thornton,
Joanna Tess,
Subhash Kumar Kolar
}
\IEEEauthorblockA{
\textit{University of Illinois Chicago} \\
Chicago, IL, USA \\
\{bkarac3, bdieugen, jtess2, skolar\}@uic.edu, patrickcnm@gmail.com
}
}

\maketitle

\begin{abstract}
Clinical framing—the linguistic manner in which clinical information is presented—can influence patient understanding and decision-making, with important implications for healthcare outcomes. Obstetrics is a high-stakes domain in which physicians counsel patients on delivery mode choices such as vaginal birth after cesarean (VBAC) and repeat cesarean section (RCS), yet counseling language remains underexplored in large-scale clinical text analysis. In this work, we analyze physician counseling language in 2,024 obstetric history and physical narratives for a rigorously defined cohort of patients for whom both VBAC and RCS were clinically viable options. To control for confounding due to medical contraindications, we first construct a VBAC-eligible cohort using structured clinical data supplemented by a large language model (LLM)-based extraction pipeline constrained to grounded, verbatim evidence from free-text narratives. We then apply a zero-shot LLM framework to categorize counseling segments into predefined framing categories capturing how physicians linguistically present delivery options. Our analysis reveals a significant difference in counseling framing distributions between VBAC and RCS notes; risk-focused language accounts for a substantially larger share of counseling segments in RCS documentation than in VBAC, with category-level differences confirmed by statistical testing, highlighting the value of controlled LLM-based framing analysis in obstetric care. 
\end{abstract}


\section{Introduction}

For women with a prior cesarean, the decision between attempting a vaginal birth after cesarean (VBAC) or undergoing a repeat cesarean section (RCS) is often preference-sensitive, involving trade-offs among maternal and neonatal risks, recovery, and future reproductive plans \cite{b1}. Because both options can be clinically appropriate for many patients, counseling plays a central role in how patients understand these trade-offs. Prior work shows that linguistic framing—how equivalent information is presented in terms of risks, benefits, or recommendations—can shape perceptions and decision-making \cite{b2,b3}. In obstetric documentation, differences in how clinicians describe VBAC and RCS may therefore reflect systematic framing patterns, motivating large-scale analysis of counseling language when both options are viable. 

Because not all patients who underwent RCS are medically eligible for VBAC, including ineligible cases would confound the linguistic comparisons by mixing notes in which VBAC counseling was not clinically appropriate. We therefore focus on a VBAC-eligible cohort—patients for whom both delivery modes were clinically viable at the time of counseling. This cohort-controlled design helps attribute observed differences in counseling language between VBAC and RCS documentation to  variation in counseling framing, rather than to medical contraindications that would preclude VBAC discussion.

Operationalizing this cohort-controlled analysis at scale requires extracting clinically relevant evidence from free-text obstetric history and physical (H\&P) narratives. Key eligibility details (e.g., prior incision type or VBAC contraindications) are often missing or incomplete in structured fields but appear in narrative documentation, motivating the use of natural language processing (NLP) methods to recover eligibility cues from the text. At the same time, counseling itself is expressed primarily in unstructured language, where framing is conveyed through subtle lexical and discourse choices that are difficult to quantify with rule-based approaches alone. We therefore use large language models (LLMs) in two complementary roles: (i) to extract grounded evidence of VBAC eligibility from narratives to support cohort construction, and (ii) to categorize counseling segments into predefined framing categories for large-scale measurement of how delivery options are presented in clinical documentation.


Motivated by the need for cohort-controlled analysis and scalable measurement of counseling language in free-text obstetric narratives, we make the following contributions:

\begin{enumerate}
    \item \textbf{VBAC-Eligible Cohort Construction:}
    We construct a cohort of patients for whom both VBAC and RCS were clinically viable at the time of counseling by combining structured eligibility criteria with LLM-based extraction of grounded, verbatim eligibility evidence from obstetric H\&P narratives, reducing confounding from VBAC contraindications.

    \item \textbf{Reliability Analysis of LLM-Based Eligibility Extraction:}
    We evaluate the reliability of LLM-based extraction under different models and prompt configurations by analyzing hallucination behavior, providing practical evidence on how prompt complexity and model scale affect source-grounded information extraction from clinical narratives.

    \item \textbf{Cohort-Controlled Framing Quantification in Counseling Documentation:}
    We apply a zero-shot LLM framework to categorize counseling segments into predefined framing categories and statistically compare framing distributions between VBAC and RCS documentation, complemented by expert review on a subset of segments.
\end{enumerate}

The paper is organized as follows. Related work is reviewed in Sec.~\ref{sec:related}. Sec.~\ref{sec:data} describes the data and cohort eligibility criteria. Sec.~\ref{sec:method} presents the proposed methodology for eligibility evidence extraction and framing analysis. Results are reported in Sec.~\ref{sec:results}. Sec.~\ref{sec:conclusion} concludes with limitations and future work.


\section{Related Work}
\label{sec:related}

Implicit framing refers to subtle differences in wording, emphasis, or presentation that can lead audiences to evaluate equivalent information differently. Classic demonstrations show that mathematically equivalent descriptions can produce systematically different judgments and choices~\cite{b4}. In healthcare, implicit framing captures the mismatch between an intention to present options neutrally and the often unintentional ways in which emphasis, omission, and evaluative wording can shape patients' perceptions and decisions~\cite{b5}. Empirical studies further suggest that such framing and recommendations have been shown to differ by patients' socioeconomic status and race/ethnicity~\cite{b6}. Obstetric decision-making provides another critical example: the choice between VBAC and RCS is preference-sensitive, yet counseling is often perceived as inadequate or directive. In a study of 68 women, Folsom et al.~\cite{b7} reported that although 18 had a calculated chance of more than 70\% for successful VBAC, only two met criteria for adequate counseling, and many participants perceived their physicians as recommending RCS. More recently, Kanjanakaew et al.~\cite{b8} emphasize that when VBAC is not explicitly presented as an option, patients may be steered toward RCS by default. Together, these findings motivate systematic examination of how VBAC and RCS options are framed in routine counseling documentation.

Clinical documentation is intended to monitor a patient's condition and support communication among members of the healthcare team~\cite{b9}. With widespread adoption of electronic health records (EHRs)—now used by more than 80\% of U.S. hospitals—these narratives have become a common source for analysis, and prior work notes benefits of the transition from paper to electronic records such as reduced prescribing and documentation errors and improved information flow~\cite{b10,b11}. EHRs include both structured fields and unstructured free-text notes; while structured fields are standardized, free text contains rich clinical detail and discussions not captured in coded fields, but its unstandardized form makes extraction and analysis difficult~\cite{b12}. Free-text notes are often information-dense and can include abbreviations and acronyms, typographical errors, inconsistencies, semantic ambiguities, and incomplete information~\cite{b13,b14}. For example, the acronym "MR" may refer to "mental retardation" but could also be interpreted as "mitral regurgitation" without sufficient context~\cite{b15}. Clinicians often organize free-text documentation through the SOAP framework (Subjective, Objective, Assessment, Plan), where patient counseling is generally recorded in the "Plan" section; however, explicit Plan sections are inconsistently labeled and may be merged under combined headings such as "Assessment and Plan", making counseling documentation difficult to analyze systematically from raw notes~\cite{b16,b17,b18,b19}. These observations motivate the use of NLP methods to interpret counseling information from heterogeneous clinical documentation.

Clinical NLP has long aimed to convert unstructured EHR narratives into structured, analyzable representations for downstream tasks such as information extraction and normalization. A large body of work has focused on identifying clinically salient mentions (e.g., problems, medications, procedures) and mapping them to standardized vocabularies, often using ontology-driven, rule-based pipelines. Representative systems include MedLEE~\cite{b20}, MetaMap~\cite{b21}, and cTAKES~\cite{b22}, which leverage lexical resources and concept inventories such as the Unified Medical Language System (UMLS)~\cite{b23} to link free-text mentions to normalized medical concepts. While these tools are effective for capturing explicit clinical concepts, they are less well-suited to characterizing counseling documentation where the signal often lies in subtle wording choices, emphasis, or recommendations that may not correspond to stable ontology concepts, and where variability in phrasing and structure can complicate reliable extraction. This motivates complementary approaches that go beyond canonical entity extraction to incorporate broader contextual and discourse cues; recent review work argues that large language models (LLMs) are well-suited to analyzing complex medical texts, with potential to transform a range of healthcare applications~\cite{b24}.

Recent studies have leveraged LLMs on core clinical NLP tasks such as section identification and biomedical information extraction. For example, Zhou et al.~\cite{b25} evaluate LLMs for clinical section identification and report that providing a modest number of task examples can substantially improve performance, highlighting the capabilities of LLMs to operate effectively via prompting or in-context examples. For biomedical information extraction, recent studies show that prompt design and example selection can materially affect clinical NER performance, and that augmenting prompts with domain resources such as UMLS can further improve results~\cite{b26}. At the same time, LLM use in clinical settings raises reliability concerns—most notably hallucination, where models generate fluent but incorrect information—which is especially consequential when studying counseling language because fabricated risks, benefits, or modifiers would distort the documented content of physician–patient communication~\cite{b27}.

Taken together, prior work establishes that framing influences clinical decision-making and that EHR narratives contain the counseling evidence needed to study such effects, but extracting and interpreting this evidence remains challenging due to the variability of free-text documentation. While LLMs offer flexible context-sensitive analysis, relability concerns such as hallucination motivate approaches that remain grounded in the source documentation when analyzing counseling language.


\section{Data}
\label{sec:data}
Our dataset comprises 2,024 obstetrics history and physical (H\&P) narratives from patients who underwent either repeat cesarean section (RCS) or vaginal birth after cesarean (VBAC). Notes were collected and managed in Research Electronic Data Capture (REDCap)—a secure web application for building and managing online surveys and databases~\cite{b28,b29}—and handled within a secure, HIPAA (U.S. health data privacy law)-compliant research environment.

Notes were drawn retrospectively from the electronic health records of a large academic medical center in the United States over a multi-year period. All available obstetric H\&P narratives meeting the inclusion criteria were included, resulting in a consecutive, corpus-based sample rather than prospective recruitment or convenience sampling.

\subsection{Automated Deidentification Framework}
We de-identified all obstetric H\&P narratives using a hybrid workflow that combines automated PHI masking with manual verification to enable privacy-preserving linguistic analysis. Automated de-identification was implemented in Spark NLP framework using pretrained clinical models augmented with targeted regular-expression rules to improve coverage for obstetric documentation~\cite{b30}.

The pipeline performs sentence splitting and tokenization, applies a named entity recognition (NER) model trained on the n2c2 2014~\cite{b31} de-identification task to detect common PHI categories including \textit{NAME}, \textit{LOCATION}, \textit{PROFESSION}, \textit{CONTACT}, \textit{ID}, and normalizes detected spans prior to masking. We added custom patterns for provider credentials (e.g., MD/APRN/CNM)\footnote{MD: Doctor of Medicine; APRN: Advanced Practice Registered Nurse; CNM: Certified Nurse Midwife;}, dates (including dates of birth), and single-letter initials that are frequently observed in clinical narratives. All detected PHI was replaced with fixed-format placeholders (e.g., \textit{"NAME"}, \textit{"DOB"} or \textit{"INITIAL"}) to minimize re-identification risk while preserving the surrounding context needed for downstream NLP.


Following automated processing, all notes were manually reviewed to verify PHI removal. A representative subset of 100 notes—Obstetrics Notes Collection (ONC)~\cite{b32}—underwent additional quality assurance and independent review by an institutional HIPAA privacy office prior to external sharing. The remaining corpus was kept access-restricted and processed only within a HIPAA-compliant secure research environment.

\subsection[Cohort Identification and Eligibility Criteria]{Cohort Identification and Eligibility Criteria~\footnote{\footnotesize{This section is developed in consultation with a clinical domain expert.}}}
After de-identification, we constructed a clinically appropriate cohort for comparing counseling language between vaginal birth after cesarean (VBAC) and repeat cesarean section (RCS) cases. Because not all patients who ultimately underwent RCS were medically eligible to attempt VBAC, including ineligible RCS cases would confound linguistic comparisons by introducing notes in which VBAC counseling was not clinically indicated. We therefore restrict analysis to patients who were VBAC-eligible at the time of counseling—i.e., those for whom either delivery mode was a viable clinical option—so that observed language differences more plausibly reflect counseling framing rather than underlying medical contraindications.

In addition to free-text narratives, the dataset includes structured attributes (e.g., \textit{prior\_cesarean}, \textit{mode\_of\_birth}, \textit{age}, \textit{body mass index (bmi)}) used for initial cohort filtering. Because both VBAC and RCS require at least one prior cesarean delivery, we first selected records with \textit{prior\_cesarean} marked true. We excluded 186 patients without a prior cesarean record, 5 patients with unspecified \textit{mode\_of\_birth}, and 2 duplicate entries. After these exclusions, 1,831 patients remained: 1,330 who underwent RCS and 501 who underwent VBAC.

Given that cesarean delivery has no known absolute contraindications aside from institutional or equipment limitations~\cite{b33}, we treat all VBAC patients as eligible to undergo RCS and assume counseling could reasonably include discussion of both delivery options prior to the patient's decision. In contrast, patients who underwent RCS may have had contraindications to VBAC, which could lead clinicians to omit VBAC from counseling discussions. To ensure that VBAC-RCS language comparisons reflect counseling among patients for whom both modes were viable, we further examined the 1,330 RCS cases and excluded those with documented contraindications to VBAC.

Eligibility criteria were derived from clinical guidelines and supporting evidence. Per the American College of Obstetricians and Gynecologists (ACOG), patients at high risk of uterine rupture—such as those with prior classical (vertical) or T-shaped uterine incisions, a prior uterine rupture, or extensive surgery involving the upper uterus—and those with contraindications to vaginal delivery (e.g., placenta previa) are generally not candidates for VBAC~\cite{b34}. Prior work also identifies interdelivery interval as a key risk factor; in particular, an interval of less than 18 months is associated with increased uterine rupture risk~\cite{b35}. At the same time, ACOG notes that women with two prior low-transverse cesarean deliveries may be reasonable VBAC candidates under appropriate conditions~\cite{b34}. Beyond absolute contraindications, prior successful vaginal birth has been reported as a protective factor associated with lower uterine rupture risk~\cite{b36}. To operationalize these criteria, we used a structured pregnancy-history table linked to each note via \textit{record\_id} to summarize prior deliveries and compute the interval between the current and most recent delivery.

Using these structured indicators and guideline-informed criteria, we categorized the 1,330 RCS patients into five VBAC eligibility groups: \textit{Eligible}, \textit{Potentially Eligible}, \textit{Limited Eligibility}, \textit{Contraindicated}, and \textit{Unknown}. We define these categories as follows:

\begin{itemize}
\item \textbf{Eligible:} Patients with one or two prior cesarean deliveries, all documented as low-transverse incisions; or patients with more than two prior cesareans but a history of successful VBAC. We also included a small subset with unknown incision type when there was evidence of prior successful vaginal birth, which is strongly suggestive of low-transverse incision. In all cases, the interdelivery interval was $\geq$ 18 months.
\item \textbf{Potentially Eligible:} Patients with one or two prior cesarean deliveries where at least one incision type was undocumented/unknown, and interdelivery interval $\geq$ 18 months.
\item \textbf{Limited Eligibility:} Patients without absolute contraindications but with clinically non-ideal factors for VBAC, such as $>$2 prior cesareans, interdelivery interval $<$ 18 months, or both.
\item \textbf{Contraindicated:} Patients with prior classical (vertical), T-shaped, or J-shaped uterine incision types, which constitute absolute contraindications to VBAC.
\item \textbf{Unknown:} Patients lacking pregnancy-history data in the structured dataset.
\end{itemize}

\begin{figure}[t]
\centering
\includegraphics[width=0.45\textwidth]{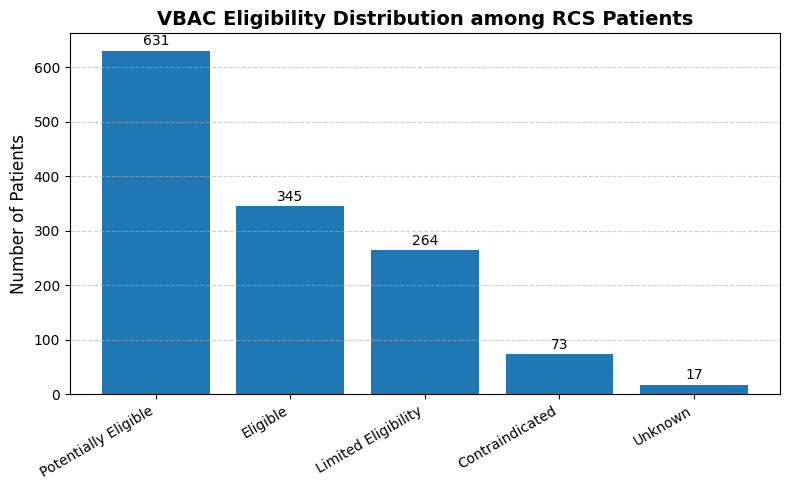}
\caption{Distribution of VBAC eligibility categories among 1,330 RCS patients.}
\label{fig:vbac_eligibility_distribution}
\end{figure}

Fig.~\ref{fig:vbac_eligibility_distribution} summarizes the distribution of these eligibility categories among RCS patients: \textit{Potentially Eligible} (47.4\%), \textit{Eligible} (25.9\%), \textit{Limited Eligibility} (19.9\%), \textit{Contraindicated} (5.5\%), and \textit{Unknown} (1.3\%).

To further characterize underlying clinical drivers of eligibility, Table~\ref{tab:vbac_conditions} reports the distribution of key condition types—number of prior cesareans, incision type, and interdelivery interval—within each eligibility category (excluding \textit{Unknown}). This breakdown illustrates how eligibility status was derived and highlights heterogeneity in risk profiles across RCS patients.


\begin{table}[t]
\caption{Counts of eligibility-related conditions by VBAC eligibility category (RCS; N=1{,}330). 
Elig.=Eligible; Pot.=Potentially eligible; Lim.=Limited eligibility; Contra.=Contraindicated.}

\label{tab:vbac_conditions}
\centering
\renewcommand{\arraystretch}{1.25}
\setlength{\tabcolsep}{4pt}
\begin{tabular}{lcccc}
\toprule
\textbf{Condition type} &
\textbf{Elig.} &
\textbf{Pot.} &
\textbf{Lim.} &
\textbf{Contra.} \\
\midrule
1 prior CS & 241 & 379 & 34 & 36 \\
2 prior CS & 95 & 252 & 30 & 23 \\
$\geq$3 prior CS & 9 & 0 & 200 & 14 \\
\midrule
Low-transverse incision & 336 & 0 & 51 & 0 \\
Unknown incision & 4 & 597 & 153 & 0 \\
Low-transverse + unknown & 5 & 34 & 60 & 0 \\
\midrule
Classical/T/J incision & 0 & 0 & 0 & 73 \\
Interdelivery interval $<$18 mo & 0 & 0 & 89 & 8 \\
Interdelivery interval $\geq$18 mo & 345 & 631 & 175 & 65 \\
\bottomrule
\end{tabular}
\end{table}


\section{Methodology}
\label{sec:method}

In this section, we first use LLMs to recover missing VBAC eligibility evidence from free-text H\&P narratives for \textit{Potentially Eligible} patients, and then analyze counseling-related excerpts from H\&P narratives to quantify cohort-level differences in framing between VBAC and RCS.
\subsection{LLM-based Extraction of Eligibility Information from H\&P Narratives in Potentially Eligible Patients}

While structured data enabled eligibility categorization for most RCS patients, a substantial subset (47.4\%) fell into the \textit{Potentially Eligible} group due to missing incision type information. Patients labeled \textit{Contraindicated} or \textit{Unknown} were excluded from further analysis because they either had absolute contraindications to VBAC or lacked sufficient pregnancy-history data for meaningful assessment. We also excluded \textit{Limited Eligibility} cases (e.g., $>$2 prior cesarean deliveries and/or interdelivery intervals $<$18 months), as these factors represent clinically significant risk conditions under standard guidelines and would complicate comparisons among patients for whom both VBAC and RCS were viable options. Consequently, we focused on \textit{Potentially Eligible} cases and attempted to recover missing eligibility evidence from the free-text H\&P narratives. Specifically, we used LLMs to extract cues related to \textit{incision types}, \textit{contraindications}, and \textit{previous delivery modes}.

\subsubsection{Prompt configurations and extraction schema}
Prior work suggests that prompt length can affect extraction accuracy and efficiency: longer prompts may improve performance in nuanced clinical tasks~\cite{b37} but increase computational cost, while prompt reduction can improve efficiency without degrading quality~\cite{b38}. Guided by these findings, we designed two instruction-based, schema-constrained prompt configurations: a concise \textit{short} prompt and a context-rich \textit{long} prompt (Fig.~\ref{fig:prompts}). Both configurations require a machine-readable JSON output schema with three fields: \textit{incision\_types}, \textit{contraindications}, and \textit{previous\_delivery\_modes}. The \textit{short} prompt provides a compact, schema-constrained formulation intended for efficient large-scale inference. The \textit{long} prompt builds on this base by adding hierarchical task decomposition, explicit examples, and repeated verbatim-only constraints intended to reduce unsupported generation.


We evaluated both prompts using two instruction-tuned Llama models: \textit{Llama 3.1 8B Instruct} and \textit{Llama 3.3 70B Instruct}. For each patient, the full H\&P narrative was provided as input, and the model output was saved as JSON for downstream validation and eligibility adjudication.

\subsubsection{Hallucination auditing: automated flags and manual verification}
Although outputs were schema-constrained, LLMs may still generate content not present in the source narrative. To assess grounding, we conducted a hallucination analysis for each extracted field. We first applied an automated verbatim-matching procedure: both the extracted strings and the corresponding H\&P narrative were normalized (case-folded, whitespace-cleaned, punctuation-stripped), and any extracted element not found verbatim in the normalized note was flagged. Flagged items were logged per patient into a JSON file under \textit{hallucinated\_incision\_types}, \textit{hallucinated\_contraindications}, and \textit{hallucinated\_previous\_delivery\_modes} to enable qualitative and quantitative analysis by model and prompt type.

Automated verbatim matching can over-flag outputs when the model produces meaning-preserving variants (e.g., abbreviation expansion) or corrects typographical errors in the source note, even under explicit "verbatim-only" instructions. To distinguish true hallucinations from benign deviations, we manually reviewed all flagged items and assigned each to one of the following six categories:

\begin{itemize}
    \item \textbf{No hallucination—semantically accurate paraphrase:} 
    The extracted content preserves the original meaning but differs lexically from the source text (e.g., ``hx'' $\rightarrow$ ``history'').

    \item \textbf{No hallucination—typo in original:} 
    The model corrects a misspelling or typographical error present in the source note.

    \item \textbf{No hallucination—typo in generated output:} 
    The extracted content matches the source evidence but contains a minor typographical error introduced by the model.

    \item \textbf{Unsupported addition (correct but non-verbatim):} 
    The model inserts a plausible absence or normalization statement (e.g., ``incision type not specified'') that is factually consistent with the note but not explicitly stated in the source text.

    \item \textbf{Hallucination:} 
    The model generates content for which no supporting evidence exists in the source narrative.

    \item \textbf{Partial hallucination:} 
    The extracted output contains a mixture of supported and unsupported information, such as correctly identifying one element while fabricating another.
\end{itemize}

For model-level comparisons, to be discussed in Sec.~\ref{sec:cohort_finalization}, these fine-grained categories were further aggregated into three broader outcome groups: \textbf{verbatim match} (exact substring present in the note), \textbf{no-hallucination variants} (meaning-preserving but non-verbatim outputs), and \textbf{hallucination variants} (unsupported content, including partial fabrications or unverifiable additions). For downstream cohort adjudication, we retained only outputs supported by the source narrative (verbatim matches and manually verified no-hallucination variants) and discarded hallucination variants.



\subsection{LLM-based Framing Analysis of Counseling Segments}
After recovering missing eligibility evidence to define an analytic cohort in which VBAC and RCS are clinically viable options, we examine counseling language in H\&P narratives to quantify framing differences between patients who ultimately delivered via VBAC versus RCS. We focus on counseling-related portions of H\&P narratives, i.e., sections where clinicians typically document delivery planning and discussion of VBAC versus RCS (e.g., assessment/plan-style sections). We characterize the \textit{tone} and \textit{style} of these excerpts to capture persuasive and informational strategies used when presenting delivery options.

\subsubsection{Framing Framework}
Prior work in health communication shows that equivalent medical information can be evaluated differently depending on how it is presented (i.e., \textit{framing effects}) \cite{b3}. Building on this literature, we operationalize counseling framing using seven categories that capture common persuasive and informational strategies observed in clinician--patient communication. Below we briefly define each category in the context of VBAC versus RCS counseling.

\begin{itemize}
    \item \textbf{Risk-Focused:} Loss-framed counseling that emphasizes complications, adverse events, or negative consequences associated with a delivery option \cite{b39,b40}. In obstetric notes, this appears as explicit emphasis on risks (e.g., uterine rupture, hemorrhage, infection) that may shape perceived safety.

    \item \textbf{Benefit-Focused:} Gain-framed counseling that emphasizes advantages or positive outcomes associated with a delivery option \cite{b41,b39}. In this setting, clinicians highlight potential benefits such as improved recovery or avoidance of major surgery.

    \item \textbf{Directive:} A recommendation-oriented style where the clinician strongly steers the patient toward a specific option, reflecting a more paternalistic interaction model \cite{b42}. This framing is characterized by authoritative wording (e.g., ``strongly recommend'') with limited collaborative language.

    \item \textbf{Shared Decision-Making:} Collaborative counseling language that explicitly involves the patient in deliberation and choice, consistent with shared decision-making models \cite{b43,b44}. Examples include offering alternatives, eliciting preferences, and emphasizing patient agency.

    \item \textbf{Balanced Information:} Neutral presentation of risks and benefits for multiple options without privileging one, supporting informed choice and mitigating biased emphasis \cite{b45}. In notes, this appears as side-by-side discussion of pros/cons for VBAC and RCS.

    \item \textbf{Statistical Evidence:} Use of numerical information (probabilities, percentages, frequencies) to frame outcomes and support decision-making \cite{b46}. In obstetric counseling, this includes statements such as predicted VBAC success rates or quantified complication risks.

    \item \textbf{Reassuring:} Emotionally supportive language intended to reduce anxiety and build trust (e.g., reassurance about monitoring or emergency preparedness) \cite{b47}. This framing focuses on empathy and comfort rather than primarily conveying factual tradeoffs.
\end{itemize}

\subsubsection{Framing Classification}
We performed segment-level framing classification using \textit{Llama 3.3 70B Instruct} in a zero-shot setup. Segments correspond to short contiguous excerpts within counseling-related portions of H\&P narratives. For each segment, the model received (i) a brief role instruction (clinical note analysis), (ii) the definitions of the seven framing categories, and (iii) an additional label, \textit{Not Counseling}, for excerpts without VBAC/RCS counseling or decision language. The model was constrained to produce a \textit{single} best-fitting label per segment to operationalize the dominant framing cue in a scalable and consistent way, yielding a baseline distribution of primary framing styles.

To support consistent parsing and downstream analysis, we enforced a structured output schema requiring two fields: \textit{category} (one of the predefined labels) and \textit{rationale} (1--2 sentences grounded in the excerpt). This design follows established prompting practices for improving zero-shot reliability, combining role specification, explicit task framing, and format-constrained outputs \cite{b48,b49,b50}. \textit{Not Counseling} segments were excluded from subsequent framing distribution and cohort-comparison analyses. Our primary analysis compares the distribution of framing categories between VBAC and RCS counseling segments; we assess association using a Pearson chi-squared test, with residual-based interpretation reported in Sec.~\ref{sec:results}.

The primary confounder in comparing VBAC and RCS counseling language is medical eligibility for VBAC, which we control by restricting analysis to patients for whom both delivery options were clinically viable at the time of counseling. Because multiple counseling segments originate from the same note and patient, results are interpreted as descriptive documentation patterns rather than independent patient-level effects.

\begin{figure}[t]
\centering

\begin{subfigure}{\columnwidth}
\begin{tcolorbox}
\footnotesize
\textbf{Short prompt (compact).}\\
\textbf{Targets:} incision type; contraindications; prior delivery modes (past only).\\
\textbf{Constraints:} verbatim sentences only; no paraphrase/inference; if absent/uncertain $\rightarrow$ \texttt{null}; JSON only.\\
\textbf{Schema:}
\begin{verbatim}
{"incision_types": [...]/null,
 "contraindications": [...]/null,
 "previous_delivery_modes": [...]/null}
\end{verbatim}
\end{tcolorbox}
\caption{}
\end{subfigure}

\vspace{0.4em}

\begin{subfigure}{\columnwidth}
\begin{tcolorbox}
\footnotesize
\textbf{Long prompt (context-rich).}\\
\textbf{Same schema and constraints as (a), plus:}
\begin{itemize}\setlength\itemsep{0pt}\setlength\parsep{0pt}
\item Target-wise task decomposition (Incision / Contraindications / Prior delivery).
\item Examples of acceptable mentions (incision types; delivery-mode synonyms).
\item Repeated verbatim-only reminders to reduce unsupported generation.
\end{itemize}
\end{tcolorbox}
\caption{}
\end{subfigure}

\caption{Prompt configurations for eligibility evidence extraction: (a) short prompt; (b) long prompt. Both enforce the same JSON schema and verbatim-only constraint; the long prompt adds task decomposition, exemplars, and repeated constraints.}
\label{fig:prompts}
\end{figure}

\section{Results}
\label{sec:results}

In this section, we present findings on eligibility evidence recovery/cohort finalization and on expert-validated framing differences between VBAC and RCS counseling notes.

\subsection{Eligibility Evidence Recovery and Cohort Finalization}
\label{sec:cohort_finalization}
We first report results for recovering missing VBAC eligibility evidence from free-text H\&P narratives in the \textit{Potentially Eligible} subset (i.e., cases lacking structured incision-type data). We compare model–prompt configurations on the faithfulness of extracted eligibility cues and then describe how these extractions were used to finalize a clinically consistent analytic cohort for downstream counseling-framing analyses.

\begin{figure}[t]
    \centering
    \includegraphics[width=0.45\textwidth]{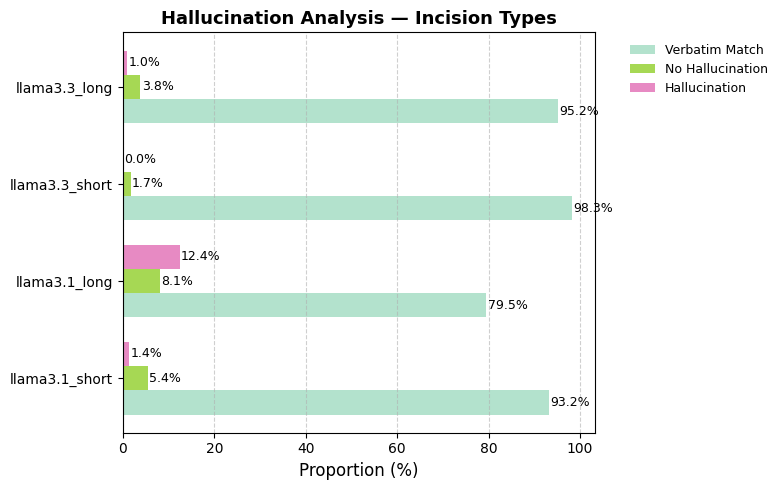}
    \caption{Distribution of output categories across models for the \textit{incision\_types} field.}
    \label{fig:hallucination_incision}
\end{figure}

\begin{figure}[t]
    \centering
    \includegraphics[width=0.45\textwidth]{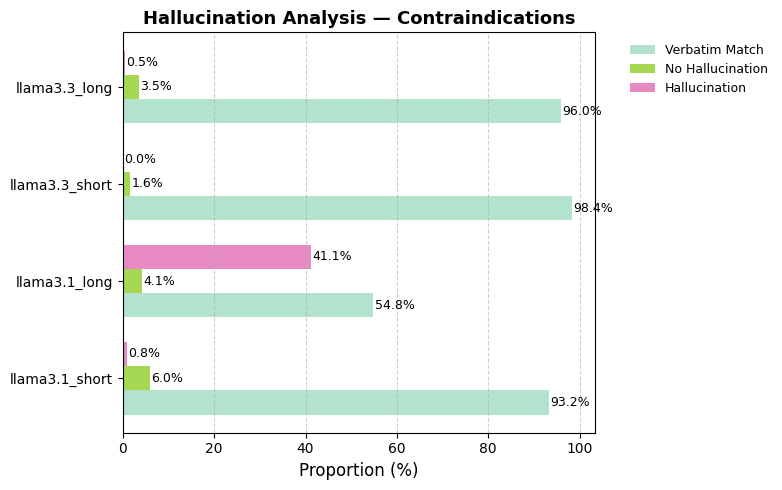}
    \caption{Distribution of output categories across models for the \textit{contraindications} field.}
    \label{fig:hallucination_contra}
\end{figure}

\begin{figure}[t]
    \centering
    \includegraphics[width=0.45\textwidth]{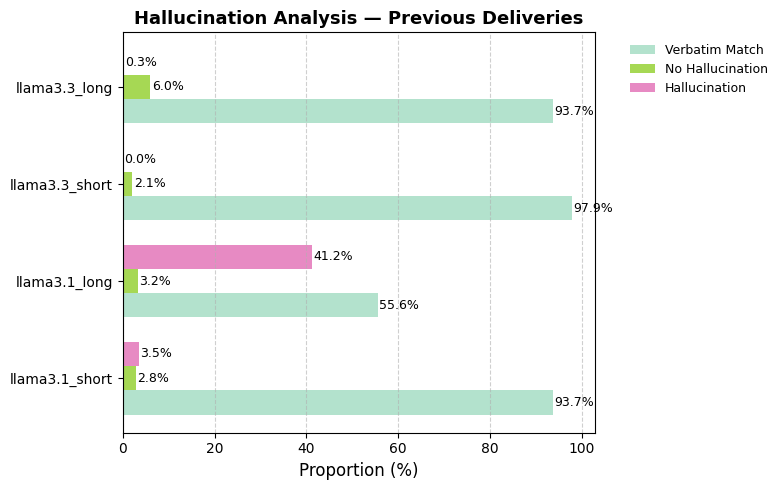}
    \caption{Distribution of output categories across models for the \textit{previous\_delivery\_modes} field.}
    \label{fig:hallucination_previous}
\end{figure}

Figs~\ref{fig:hallucination_incision}--\ref{fig:hallucination_previous} summarize extraction faithfulness for each eligibility field (\textit{incision\_types}, \textit{contraindications}, \textit{previous\_delivery\_modes}) by reporting three outcome groups: (i) \textit{verbatim matches} to the source narrative, (ii) \textit{non-hallucinated variants} (meaning-preserving deviations including paraphrases or typo corrections), and (iii) \textit{hallucination variants} (unsupported additions, including partial fabrications). This view emphasizes whether a configuration produces \emph{precise, source-grounded} extractions versus plausible-but-nonverbatim or unsupported content.

Across all three fields, \textit{Llama 3.3 70B} produced more reliable, source-grounded outputs than \textit{Llama 3.1 8B}, with performance largely robust to prompt length. In contrast, \textit{Llama 3.1 8B} was highly sensitive to prompt complexity: under the long prompt, hallucination variants exceeded 40\% for \textit{contraindications} and \textit{previous\_delivery\_modes}, whereas its short-prompt behavior was substantially more stable. Notably, \textbf{Llama 3.3 70B with the short prompt} produced \textbf{zero hallucination variants across all fields}, indicating consistently grounded extraction of eligibility evidence from the narratives. Overall, these results suggest that instruction-heavy prompts can degrade factual consistency for smaller models, while concise prompts yield more dependable extractions in clinical free text.

Table~\ref{tab:hallucination_detailed} provides a more granular breakdown of initially flagged outputs. \textit{Llama 3.3 70B} deviations were dominated by benign nonverbatim behavior (paraphrases and typo-related variants) with negligible hallucinations, whereas \textit{Llama 3.1 8B} under the long prompt shifted toward full hallucinations for \textit{contraindications} and \textit{previous\_delivery\_modes}, consistent with an instruction-overload failure mode. The short-prompt \textit{Llama 3.1 8B} configuration additionally produced a small number of “correct but extra information” cases (unsupported absence statements), suggesting that even when smaller models remain mostly stable, their errors may manifest as plausible but non-grounded additions.

Based on these faithfulness results, we used outputs from \textit{Llama 3.3 70B (short prompt)} as the primary evidence source for eligibility adjudication, consulting other configurations only when the primary output was incomplete and retaining information only when it was verified as supported. Eligibility adjudication followed guideline-based rules (e.g., treating low-transverse evidence or prior vaginal birth as supportive of eligibility and excluding classical/T/J incisions or contraindications to vaginal delivery) with normalization of common synonyms/abbreviations (e.g., Pfannenstiel, LTCS). Ambiguous surgical histories (e.g., prior myomectomy\footnote{Myomectomy is a surgical procedure to remove uterine fibroids.} or undocumented scar type) were resolved via consultation with a clinical expert. Using this process, we confirmed that \textbf{512 of 631} \textit{Potentially Eligible} patients met VBAC eligibility criteria, with remaining cases excluded due to contraindications or insufficient documented evidence. Incorporating these recovered cases with patients already deemed eligible from structured data yielded a finalized analytic cohort of \textbf{1,358} patients: \textbf{501 VBAC} and \textbf{857 RCS} (\textit{512 confirmed from the potentially eligible subset} and \textit{345 previously eligible}). This cohort construction step enabled downstream analyses on counseling content in a population where both VBAC and RCS were clinically viable options.

\begin{table}[t]
\centering
\caption{Distribution (\%) of initially flagged error types by model--prompt and extraction field.}
\label{tab:hallucination_detailed}
\scriptsize
\setlength{\tabcolsep}{2.7pt} 
\renewcommand{\arraystretch}{1.1}
\begin{tabular}{llrrrrrr}
\toprule
\textbf{Model--Prompt} & \textbf{Field} &
\multicolumn{6}{c}{\textbf{Error type (\%)}} \\
\cmidrule(lr){3-8}
& & \textbf{Para} & \textbf{Typo-O} & \textbf{Typo-G} & \textbf{Extra} & \textbf{Part.} & \textbf{Hall.} \\
\midrule
Llama 3.1 Short & Incision & 72.1 & 7.0 & 0.0 & \textbf{7.0} & 0.0 & 13.9 \\
Llama 3.1 Long  & Incision & 35.6 & 3.9 & 0.0 & 0.0 & 3.1 & \textbf{57.4} \\
Llama 3.3 Short & Incision & \textbf{81.8} & \textbf{18.2} & 0.0 & 0.0 & 0.0 & 0.0 \\
Llama 3.3 Long  & Incision & 70.0 & 10.0 & 0.0 & 0.0 & \textbf{10.0} & 10.0 \\
\midrule
Llama 3.1 Short & Contra. & 74.4 & 14.0 & 0.0 & \textbf{7.0} & \textbf{2.3} & 2.3 \\
Llama 3.1 Long  & Contra. & 7.4 & 1.7 & 0.0 & 0.0 & 0.0 & \textbf{90.9} \\
Llama 3.3 Short & Contra. & \textbf{80.0} & 20.0 & 0.0 & 0.0 & 0.0 & 0.0 \\
Llama 3.3 Long  & Contra. & 48.0 & \textbf{32.0} & \textbf{8.0} & 0.0 & 0.0 & 12.0 \\
\midrule
Llama 3.1 Short & Prev. del. & 45.0 & 0.0 & 0.0 & 0.0 & \textbf{2.5} & 52.5 \\
Llama 3.1 Long  & Prev. del. & 6.0 & 0.4 & 0.8 & 0.0 & 0.0 & \textbf{92.8} \\
Llama 3.3 Short & Prev. del. & \textbf{100.0} & 0.0 & 0.0 & 0.0 & 0.0 & 0.0 \\
Llama 3.3 Long  & Prev. del. & 92.5 & \textbf{2.5} & 0.0 & 0.0 & \textbf{2.5} & 2.5 \\
\bottomrule
\end{tabular}
\end{table}

\subsection{Framing Analysis Validation and Cohort Comparisons}
Within the finalized cohort, we assess framing label reliability via expert review and then compare framing category distributions between VBAC and RCS counseling using both descriptive summaries and statistical tests.

\subsubsection{Expert Validation and Interpretation Considerations}
To assess whether the LLM’s framing labels align with clinical interpretation, a domain expert independently reviewed 50 counseling segments (25 VBAC, 25 RCS) sampled to ensure coverage across categories (random samples plus targeted inclusion of frequent and rare categories). The expert emphasized that some categories are intrinsically difficult to adjudicate from short, decontextualized excerpts—especially \textit{Balanced Information}, \textit{Shared Decision-Making}, and \textit{Directive} language—because these framings often depend on conversational context (e.g., whether alternatives were genuinely offered, whether a recommendation followed patient preference elicitation, or whether the clinician’s intent was supportive versus steering). In contrast, \textit{Risk-Focused}, \textit{Benefit-Focused}, and \textit{Statistical Evidence} were judged easier to identify consistently due to more explicit lexical cues (complications, advantages, numeric probabilities).

The expert also noted that some excerpts reflect standardized “canned” documentation (e.g., “risks/benefits/alternatives discussed; patient understands”), which may satisfy institutional requirements but conveys limited framing signal. These segments are superficially compatible with multiple labels (often appearing neutral or “balanced”) while being semantically thin, highlighting an inherent limitation of note documentation rather than the classification procedure itself.

Quantitatively, agreement was evaluated using metrics appropriate for comparing the LLM’s single-label output to the expert’s multi-label judgments. Under an any-match criterion (LLM label counted correct if it appeared among the expert’s labels), agreement was \textit{80\%}. To capture partial overlap, the mean Jaccard similarity between the LLM label and the expert’s label set was \textit{68.1\%}. Finally, using the expert’s first-listed category as a proxy for a primary label, Cohen’s $\kappa = 0.56$, indicating moderate agreement. Together, these findings suggest that the LLM usually identifies the dominant framing cue in counseling excerpts, while disagreement concentrates in categories whose interpretation depends most on broader conversational context.

\subsubsection{Distribution of Framing Categories}
Across both cohorts, most extracted segments were labeled \textit{Not Counseling}, consistent with the fact that counseling-related headers (e.g., Assessment/Plan-style sections) often contain procedural documentation, follow-up logistics, or generic plan statements rather than VBAC/RCS deliberation. In the VBAC cohort, 722 of 3,848 segments (18.8\%) were classified into one of the seven framing categories; in the RCS cohort, 1,285 of 6,904 segments (18.6\%) were classified into these categories. All subsequent analyses exclude \textit{Not Counseling} segments and focus only on counseling-relevant segments.

\begin{figure}[t]
    \centering
    \includegraphics[width=\columnwidth]{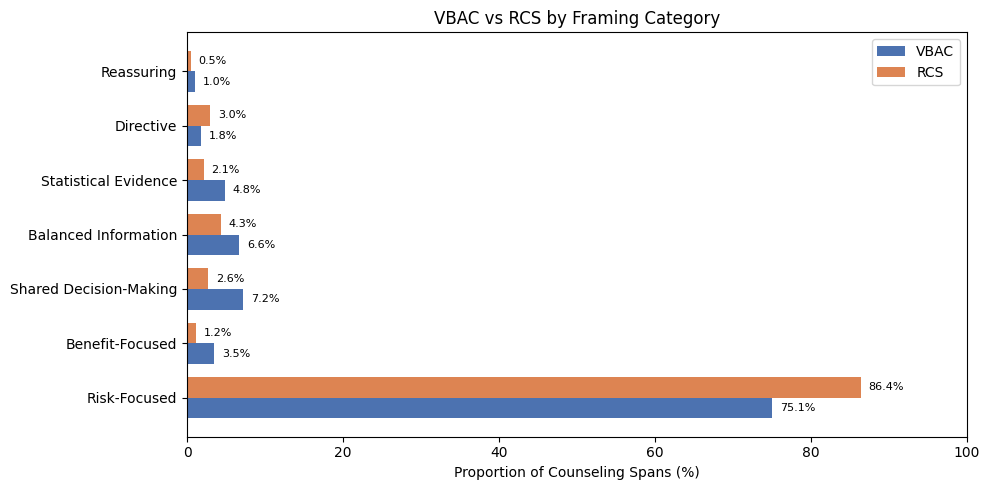}
    \caption{Comparison of framing category proportions between VBAC and RCS counseling segments.}
    \label{fig:framing_dist}
\end{figure}

Among counseling-relevant segments, framing distributions were strongly skewed toward \textit{Risk-Focused} language in both cohorts (Fig.~\ref{fig:framing_dist}). In VBAC notes, \textit{Risk-Focused} accounted for 75.1\% of counseling segments, followed by \textit{Shared Decision-Making} (7.2\%), \textit{Balanced Information} (6.6\%), and \textit{Statistical Evidence} (4.8\%). The remaining categories were comparatively rare (\textit{Benefit-Focused} 3.5\%, \textit{Directive} 1.8\%, \textit{Reassuring} 1.0\%). RCS notes exhibited an even stronger concentration: \textit{Risk-Focused} comprised 86.4\% of counseling segments, with all other categories collectively accounting for less than 15\%.

To aid interpretability of the framing categories, Table~\ref{tab:framing_examples} presents representative de-identified counseling excerpts assigned to each category by the LLM classifier. These examples illustrate the linguistic patterns the model associates with each framing type (e.g., explicit risk enumeration, recommendation-oriented phrasing, or emphasis on patient choice), rather than serving as definitive or exhaustive definitions.

\begin{table}[t]
\centering
\caption{Representative counseling excerpts assigned for each LLM-assigned framing category (de-identified).}
\label{tab:framing_examples}
\footnotesize
\renewcommand{\arraystretch}{1.05}
\setlength{\tabcolsep}{4pt}
\begin{tabular}{|l|p{0.58\columnwidth}|}
\hline
\textbf{Category} & \textbf{Example Counseling Excerpt} \\
\hline
Risk-Focused & “Risks of RCS include bleeding, infection and damage to nearby organs including bladder, bowel, tubes, ovaries, uterus, baby, nerves and vessels.” \\
\hline
Benefit-Focused & “We discussed that her possibility of success will improve if she presents in active labor.” \\
\hline
Shared Decision-Making & “Discussed that if she desires a Cesarean section at any point in the labor course that she can let us know and we will proceed.” \\
\hline
Balanced Information & “We did review that risk of rupture after one C/S and 2 C/S is $<1\%$, also reviewed the benefits of successful TOLAC including avoidance of major surgery, shorter hospitalization, less postpartum complications, possible less maternal and fetal morbidity.” \\
\hline
Directive & “Discussed strong recommendation for labor in hospital and epidural anesthesia.” \\
\hline
Statistical Evidence & “Using MFMU calculator she has a 74\% sucess rate of VBAC.” \\
\hline
Reassuring & “I reassured her that this institution is well equipped to handle emergencies surrounding a trial of labor but cannot prevent all adverse outcomes.”
\\
\hline
\end{tabular}
\end{table}

\subsubsection{Comparative Analysis Between VBAC and RCS Counseling Notes}
While Fig.~\ref{fig:framing_dist} suggests descriptive differences between cohorts, we next test whether framing distributions differ statistically between VBAC and RCS outcomes. We constructed a contingency table of framing category by delivery group (Table~\ref{tab:framing_residuals}) and applied a Pearson $\chi^{2}$ test. The test indicates a significant association between framing category and delivery group ($\chi^{2}=62.38$, $p\approx 1.48\times 10^{-11}$), rejecting the null hypothesis that framing distributions are independent of outcome.

\begin{table}[h!]
\centering
\caption{Contingency table of framing categories by delivery group (adjusted Pearson residuals in parentheses).}
\label{tab:framing_residuals}
\small
\begin{tabular}{lccc}
\hline
\textbf{Category} & \textbf{RCS} & \textbf{VBAC} & \textbf{Total} \\
\hline
\textbf{Balanced Information} & 
55 (-2.31) & 
48 (2.31) & 
103 \\
\textbf{Benefit-Focused} & 
15 (-3.53) &
25 (3.53) &
40 \\
\textbf{Directive} &
38 (1.58) &
13 (-1.58) &
51 \\
\textbf{Reassuring} &
6 (-1.35) &
7 (1.35) &
13 \\
\textbf{Risk-Focused} &
1110 (6.37) &
542 (-6.37) &
1652 \\
\textbf{Shared Decision-Making} &
34 (-4.84) &
52 (4.84) &
86 \\
\textbf{Statistical Evidence} &
27 (-3.41) &
35 (3.41) &
62 \\
\hline
\textbf{Total} &
1285 & 722 & 2007 \\
\hline
\end{tabular}
\end{table}

To identify which categories contribute most to this association, we computed adjusted Pearson residuals for each cell (Table~\ref{tab:framing_residuals}). Residuals with $|r|>2$ indicate categories that occur significantly more or less often than expected under independence~\cite{b51,b52}. \textit{Risk-Focused} language is strongly overrepresented in RCS notes ($r=+6.37$), consistent with a more uniformly risk-dominant counseling profile in this cohort. In contrast, VBAC notes show significant overrepresentation of \textit{Shared Decision-Making} ($r=+4.84$) and \textit{Benefit-Focused} framing ($r=+3.53$), suggesting relatively more participatory and gain-framed counseling among patients who ultimately delivered vaginally. \textit{Statistical Evidence} ($r=+3.41$) and \textit{Balanced Information} ($r=+2.31$) are also significantly more frequent in VBAC counseling. Differences for \textit{Directive} ($|r|=1.58$) and \textit{Reassuring} ($|r|=1.35$) do not exceed conventional significance thresholds.

Overall, \textbf{risk-oriented framing dominates counseling language in both cohorts, but is significantly more pronounced in RCS notes}. Although \textit{Shared Decision-Making}, \textit{Balanced Information}, and \textit{Benefit-Focused} language occur more frequently in VBAC counseling than in RCS counseling, these strategies remain comparatively infrequent relative to risk-focused discourse, suggesting that obstetric counseling documentation is largely loss-framed even when both delivery options are clinically viable. This emphasis on complications may shape perceived safety and, through implicit framing, influence how delivery options are evaluated.

\section{Conclusion and Future Work}
\label{sec:conclusion}
We presented a cohort-controlled analysis of clinician counseling language for VBAC versus RCS decision-making in obstetric H\&P narratives. To avoid confounding from medically ineligible cases, we constructed a VBAC-eligible analytic cohort by combining structured eligibility criteria with source-grounded, verbatim LLM extraction from free text. Within this cohort, a zero-shot LLM framing framework revealed significant differences in counseling documentation between VBAC and RCS outcomes: risk-focused framing dominated both cohorts, but was substantially more concentrated in RCS notes, while VBAC notes showed comparatively higher (though still infrequent) use of shared decision-making, balanced information, benefit-focused language, and statistical evidence. These findings demonstrate how controlled cohort construction paired with LLM-based framing categorization can enable scalable measurement of counseling patterns in routine clinical documentation. By quantifying systematic differences in counseling framing under controlled eligibility conditions, this work provides a foundation for future studies examining how documentation practices may influence patient understanding and shared-deicision making when multiple clinically viable options exist.

This study also has important limitations. First, framing labels were assigned to short, contiguous excerpts using a single-label formulation, which can introduce ambiguity—particularly for categories such as \textit{Directive}, \textit{Shared Decision-Making}, and \textit{Balanced Information} that depend on broader textual context. This design intentionally prioritizes scalable measurement of the dominant framing signal, and the moderate expert agreement ($\kappa=0.56$) suggests that primary framing cues are generally captured even though mixed or context-dependent framing may be missed. Second, our analysis quantifies \textit{how} counseling is framed, but does not yet characterize \textit{what} clinical content is communicated (e.g., which risks, benefits, or patient-specific factors are discussed), limiting clinical interpretability of why certain framings differ across cohorts. Third, we use a zero-shot LLM due to limited annotations, enabling scalability but precluding comparison with supervised or hybrid methods. Fourth, the dataset is from a single center, and results are descriptive—without linking framing to patient decisions or outcomes. In addition, statistical tests treat segments as independent despite within-patient clustering.  

Future work will address these limitations in two complementary directions. (i) \textbf{Richer and more context-aware framing representations:} rather than treating counseling as isolated segments, we will incorporate broader context (neighboring sentences, section structure, or note-level counseling spans) and allow multi-label or hierarchical framing assignments to better capture mixed strategies within a counseling episode. We will also explore note-level or patient-level aggregation schemes that reduce sensitivity to local ambiguity while preserving clinically meaningful differences, and incorporate expert-annotated subsets to enable supervised or hybrid framing models. (ii) \textbf{Content-focused counseling analysis with normalization:} we will move beyond framing categories to extract the clinical \emph{substance} of counseling (e.g., specific risks, procedures, symptoms, and outcomes discussed) and standardize these mentions to unified concept identifiers. Such ontology-based normalization can reduce confounding from lexical variability (e.g., ``bleeding'' vs.\ ``hemorrhage''; ``risk of'' vs.\ ``chance of'') and enable reproducible comparisons of counseling content across clinicians and delivery groups. Together, these extensions would support more fine-grained, clinically actionable analyses of both the framing and the informational content of obstetric counseling documentation.








\end{document}